\documentclass[10pt,twocolumn,a4paper]{article}

\usepackage[british]{babel}

\usepackage[table]{xcolor} %

\usepackage[textsize=footnotesize,textwidth=1.5cm]{todonotes}
\usepackage[T1]{fontenc}
\usepackage{lmodern}

\usepackage[left=2.25cm,right=2.25cm,top=2.20cm,bottom=2.20cm]{geometry}
\usepackage{booktabs}
\usepackage{times}
\usepackage{epsfig}
\usepackage{graphicx}
\usepackage{amsmath} %
\usepackage{amssymb} %
\usepackage{amsthm} %
\usepackage{array}
\usepackage{mathtools}
\usepackage{algorithm}
\usepackage{algorithmic}
\usepackage{subfig}
\usepackage{varwidth}%
\usepackage{mdframed}
\usepackage[export]{adjustbox} 
\usepackage{dsfont}
\usepackage{xfrac}
\usepackage{pgfplots}
\pgfplotsset{compat=1.12}
\usepackage{tikz}
\usetikzlibrary{positioning}
\usetikzlibrary{calc}
\usetikzlibrary{patterns}

\usepackage[noblocks]{authblk}

\graphicspath{{./}}

\newcommand{\blkdiag}[1]{\text{blkdiag}\left( #1\right)}
\DeclareMathOperator{\codim} {codim}
\renewcommand{\vec}[1]{\text{vec}\left( #1\right)}
\newcommand{\vecemph}[1]{\emph{\text{vec}}\left( #1\right)}
\newcommand{\tr}[1]{\text{tr}\left( #1\right)}
\newcommand{\rank}[1]{\text{rank}\left( #1\right)}
\newcommand{\rankemph}[1]{\emph{\text{rank}}\left( #1\right)}
\newcolumntype{L}[1]{>{\raggedright\let\newline\\\arraybackslash\hspace{0pt}}m{#1}}
\newcolumntype{C}[1]{>{\centering\let\newline\\\arraybackslash\hspace{0pt}}m{#1}}
\newcolumntype{R}[1]{>{\raggedleft\let\newline\\\arraybackslash\hspace{0pt}}m{#1}}

\newcommand{\ones}{{\mathds{1}}}

\newcommand{\SO}[1]{{\rm SO}(#1)}    %
\newcommand{\Q}{{\mathcal Q}}    %

\newtheorem{lemma}{Lemma}
\theoremstyle{remark}
\newtheorem{example}{Example}
\theoremstyle{plain}
\newtheorem{corollary}{Corollary}

\makeatletter
\newcommand\footnoteref[1]{\protected@xdef\@thefnmark{\ref{#1}}\@footnotemark}
\makeatother

\makeatletter
\newenvironment{procedure}[1][htb]
  {
   \renewcommand*{\ALG@name}{Procedure}
   \begin{algorithm}[#1]%
  }{\end{algorithm}}
\makeatother

\newcommand{\mcell}{\cellcolor{green!25} Minimal}
\newcommand{\amcell}{\cellcolor{yellow!25} Almost min.}
\newcommand{\nmcell}{\cellcolor{red!25} Not minimal}
\newcommand{\tcell}{\cellcolor{green!25} Always tight}
\newcommand{\ntcell}{\cellcolor{yellow!25} Non-tight instances found}
\newcommand{\lncell}{\cellcolor{red!25} Low noise}

\makeatletter
\patchcmd{\@maketitle}{\LARGE}{\fontsize{16.9}{17}\selectfont}{}{}
\makeatother
\title{
\vspace{1em}
On the Tightness of Semidefinite Relaxations for Rotation Estimation%
\vspace{.7em}
}

\author[1]{
Lucas Brynte
}
\author[2]{
Viktor Larsson
}
\author[1]{
Jos\'e Pedro Iglesias
}
\author[1,3]{
Carl Olsson
}
\author[1]{
Fredrik Kahl
}
\affil[1]{
Chalmers University of Technology, Gothenburg, Sweden.
Email:~\texttt{\{brynte,jose.iglesias,caols,fredrik.kahl\}@chalmers.se}
\vspace{.5em}
}
\affil[2]{
ETH Zurich, Zurich, Switzerland.
Email:~\texttt{viktor.larsson@inf.ethz.ch}
\vspace{.5em}
}
\affil[3]{
Lund University, Lund, Sweden.
}

\date{}

\begin{document}
\maketitle

\begin{abstract}
Why is it
that semidefinite relaxations have been so successful
in numerous
applications in computer vision and robotics for solving non-convex optimization problems involving rotations? In studying the empirical
performance
we note that there are few failure cases reported in the literature, in particular for estimation problems with
a single
rotation,
motivating us
to gain further theoretical understanding.

A general framework based on tools from algebraic geometry is introduced for analyzing the power of semidefinite relaxations of problems with quadratic objective functions and rotational constraints.
Applications include registration, hand-eye calibration
and rotation averaging.
We characterize the extreme points, and show that there exist failure cases for which the relaxation is not tight, even in the case of a single rotation.
We also show that some problem classes are always tight given an appropriate parametrization. %
Our theoretical findings are accompanied with numerical
simulations, providing
further evidence and understanding of the results.
\end{abstract}

\section{Introduction}
Optimization over the special orthogonal group %
of the orthogonal matrices with determinant one occurs in many geometric vision problems where %
rigidity of a model needs to be preserved under transformations.
While the objective functions are often simple least squares costs, constraining a matrix to be a rotation requires a number of quadratic equality constraints on the elements which makes the problem non-convex.
On the other hand, since both the objective and the constraints are quadratic the Lagrange dual function can be computed (in closed form) and therefore optimization of the dual problem can be considered. It turns out that this is a linear semidefinite program (SDP), which can be reliably solved with standard solvers in polynomial time.

\begin{figure}[t!]
\begin{center}
   \setlength\tabcolsep{0pt} %
   \begin{tabular}{cc} 
   \includegraphics[width=0.5\linewidth]{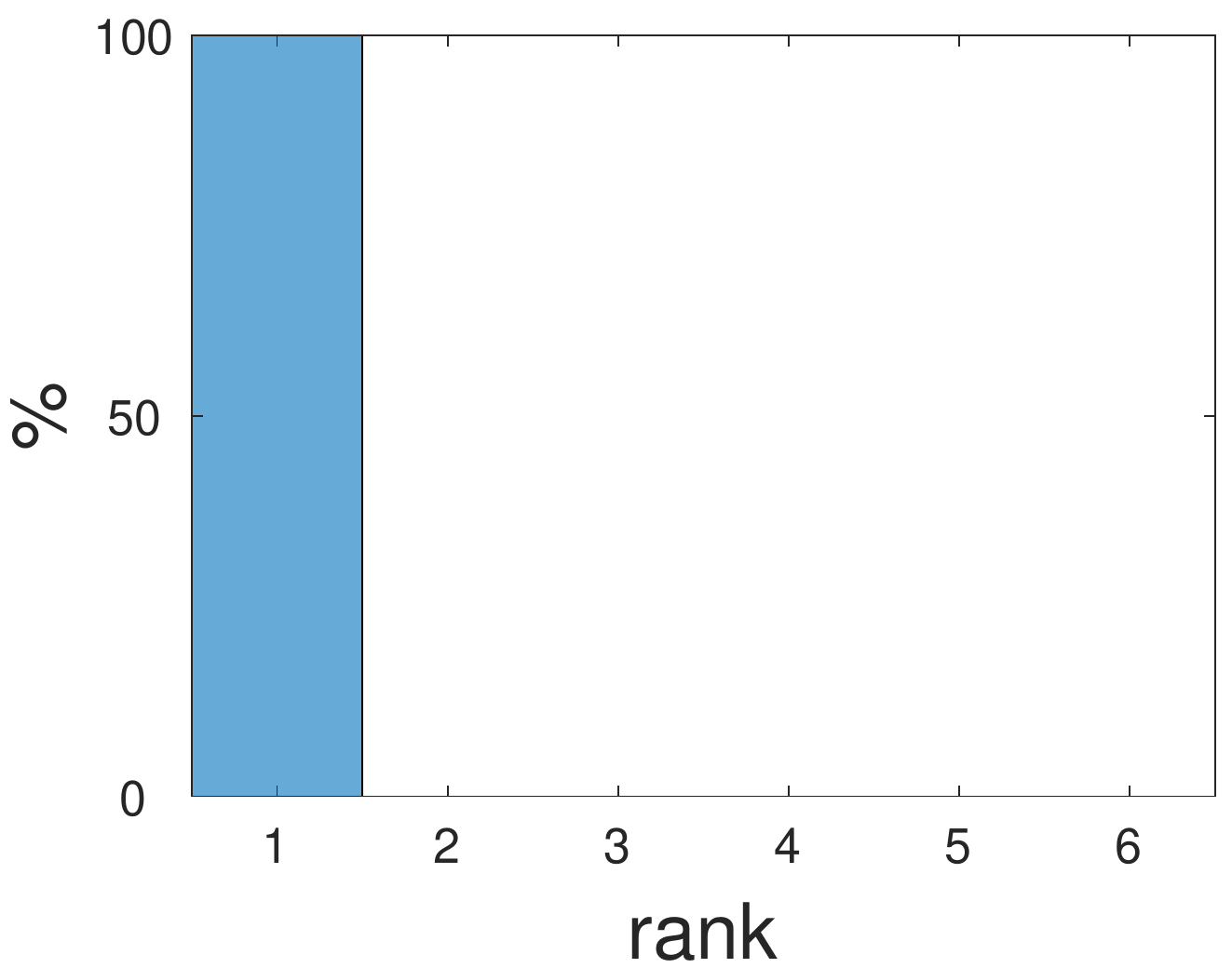} &
   \includegraphics[width=0.5\linewidth]{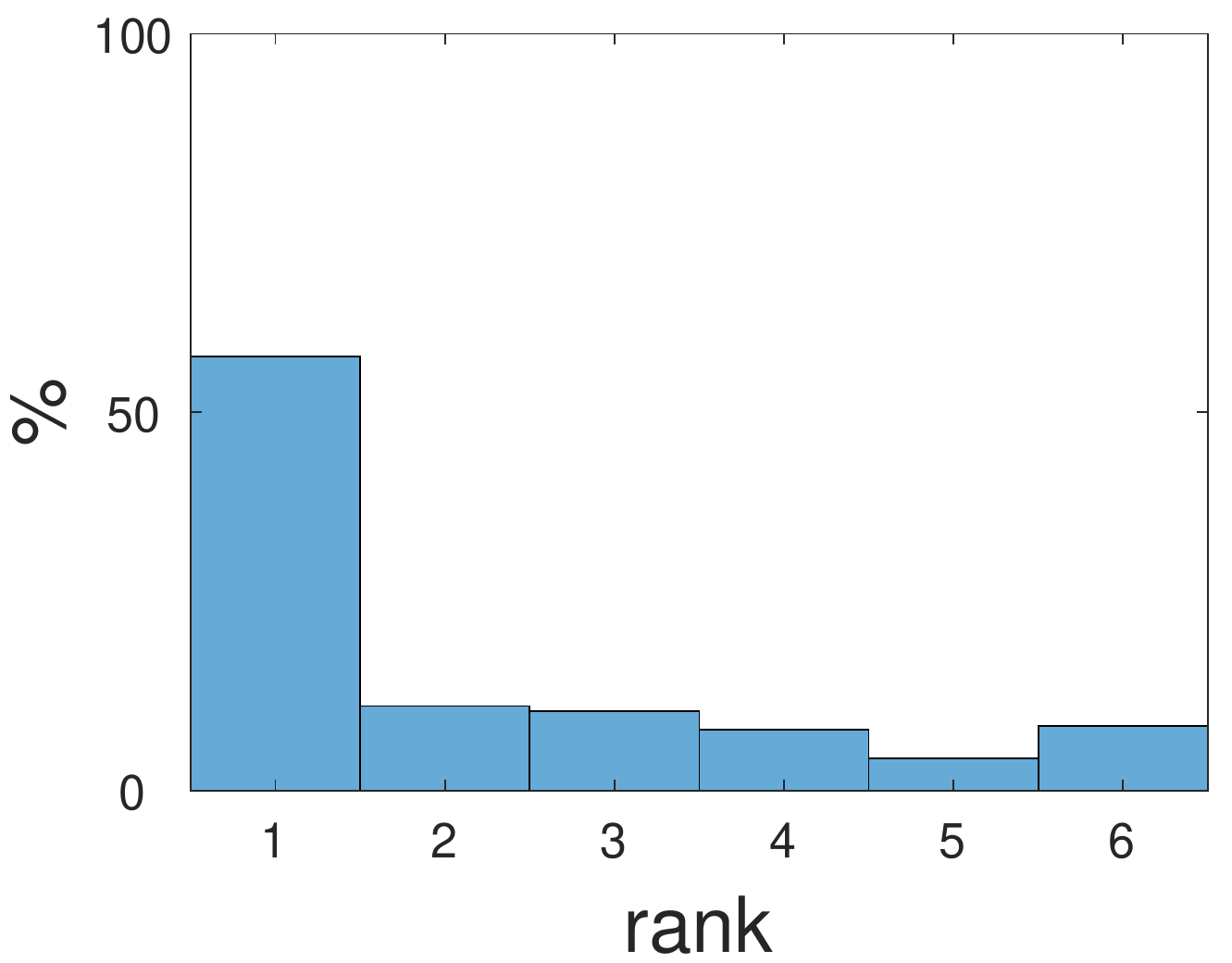} \\
   (a) & (b)
  \end{tabular} \vspace{-0.5cm}
   \setlength\tabcolsep{6pt} %
\end{center}
   \caption{Histogram of the rank of the extreme points for $1000$ synthetic experiments. If the rank %
   is one, then the globally optimal solution can be extracted. (a) Problem instances with a quadratic objective function defined over $\SO{3}$ are solved via an SDP relaxation. The coefficients for the quadratic function of each instance are uniformly drawn from $[\,\text{-}1,1\,]$. (b) The experiments are performed in a similar
   manner,
   except that the quadratic functions are defined over $\SO{3} \times \SO{3}$. %
   In almost $50\%$ of the cases, an extreme point with rank greater than one is obtained, and hence the SDP relaxations are not tight. See Section~\ref{sec:examples} for details.
   }
\label{fig:empirical1}
\end{figure}

Recent studies \cite{briales-cvpr-2017,eriksson-etal-pami-2019,iglesias-etal-cvpr-2020,carlone-etal-iros-2015,Chaudhury-etal-siam-2015,mangelson-etal-icra-2019} have observed that in many practical applications the lower bound provided by the dual problem is often the same as the optimal value of the primal one. In such cases duality offers a way of solving the original non-convex problem using a tight convex relaxation.
For different problem classes, the prevalence of problem instances with tight relaxations varies, as illustrated by the synthetically generated instances in Figure~\ref{fig:empirical1}.
Furthermore,
finding conditions that can be rigorously proven to be sufficient for a tight relaxation and strong duality for a given problem class remains a challenging research area.  
In this paper we focus on the converse question: 
For what problem classes can we find objective functions that give a non-zero duality gap and a non-tight relaxation?
We use tools from algebraic geometry for analyzing when this happens in the case of general quadratic objective functions over rotational constraints. In particular we consider the three most commonly occurring parametrizations that can be realized with quadratic constraints, namely 3D-rotations represented either by matrices from $\SO{3}$ or by unit norm quaternions, and planar rotations represented by matrices from $\SO{2}$.

We consider the dual of the dual wherein all quadratic terms of the primal problem are replaced by linear terms over a set of `lifted' variables subject to semidefinite constraints. 
The quadratic objective functions are then replaced by linear functions in the new variables which are known to attain their optimum in extreme points of the feasible set.
By studying the extreme points
of this relaxation we show that
the situation is not as favorable as one might expect from the literature:
Even for applications with relatively few rotations, we prove the existence of
extreme points with rank strictly greater than one, and objective functions which are minimized at such points.
The larger rank
then
prevents us from extracting a solution to the primal problem from the `lifted'
variables,
and shows that there is a duality gap.

Our main contributions are:
\begin{itemize}
    \item We present a novel analysis of the duality properties for quadratic minimization problems subject to rotational constraints based on algebraic geometry.
    
    \item We characterize for several applications when the standard SDP relaxation is tight and when it is not. For instance, we give counterexamples for the registration problem with $\SO{3}$-parametrization, showing that the SDP relaxation is not always tight
    since its solution may be an extreme point of rank~6.
    Similarly, we generate counterexamples which show that averaging of four planar rotations is not necessarily tight.

    \item We show that the registration problem and the hand-eye calibration problem with $\SO{2}$-parametrization or quaternion parametrization are always guaranteed to produce tight SDP-relaxations.

\end{itemize}

\subsection{Related work}

It is well-known that finding the optimal rigid transformation that registers two point clouds can be done in closed form \cite{horn-etal-1987}. This is a key subroutine used in many different applications, for example, in the ICP algorithm \cite{besl-mckay-pami-1992}. However, registering other geometric primitives is a much harder problem. In \cite{olsson-etal-cvpr-2006,olsson-etal-pami-2009}, a branch-and-bound approach is proposed for finding the 3D rigid transformation for corresponding points, lines and planes. The same problem is solved in \cite{briales-cvpr-2017} by first eliminating the translation and then using SDP relaxations for estimating the rotation. Empirically, it was noted that the relaxations were always tight, but no theoretical analysis was given. The problem of registering multiple point clouds was solved using SDP relaxations and Lagrangian duality in \cite{Chaudhury-etal-siam-2015}. The problem was further studied in \cite{iglesias-etal-cvpr-2020} where it was shown that for low noise regimes, the SDP relaxation is always tight.

In robotics, SDP relaxations for estimating rigid transformations in simultaneous localization and mapping (SLAM) have been explored in a number of recent papers~\cite{carlone-etal-iros-2015,carlone-etal-tr-2016,briales2016fast,mangelson-etal-icra-2019,fan2020}.
Again, the empirical performance %
is generally good, the optimal solutions can be efficiently computed~\cite{boumal2015riemannian}, and the relaxations are shown to be tight for bounded noise levels. Non-tight counterexamples are also provided. 
In computer vision, there are many structure-from-motion (SfM) pipelines that rely on solving the so-called rotation averaging problem, see \cite{hartley-etal-cvpr-2011,fredriksson2012simultaneous,chatterjee-govindu-iccv-2013,hartley2013,arrigoni2014robust,Wilson2016,eriksson-etal-pami-2019,dellaert-eccv-2020}. One of the first approaches to use convex relaxations and duality in this setting was \cite{fredriksson2012simultaneous} where it was empirically observed that the relaxations are tight. %
A theoretical analysis and proof that for low noise regimes, SDP relaxations are indeed tight (no duality gap) have been derived in \cite{cifuentes-etal-2017,rosen-etal-2018,eriksson-etal-pami-2019} for the problem of rotation averaging. The recent paper \cite{dellaert-eccv-2020} explores this analysis to develop an efficient algorithm with optimality guarantees.

Estimating the pose of a camera also involves optimization over the special orthogonal group, \cite{zheng2013revisiting,sweeney2014gdls,nakano2015globally,brynte-kahl-bmvc-2020}. Approaches based on minimal solvers and Gr{\"o}bner bases are often used. Alternatively, we show that the camera pose problem can be solved with SDP relaxations and convex optimization. Another classical problem that involves rotations is the hand-eye calibration problem \cite{horaud-dornaika-1995}. In a recent paper \cite{giamou-etal-2019}, an SDP relaxation is proposed, again, with seemingly good empirical performance.

There are several previous works with similar aims as ours, but for more general problem settings.
For a general, geometric overview of the problem at hand, we recommend \cite{sanyal_sottile_sturmfels_2011} where orbitopes are studied. An orbitope is the convex hull of an orbit of a compact group acting linearly on a vector space. The dual of the dual formulation that we study corresponds to the first order relaxation of the moment-SOS hierarchy \cite{henrion2020momentsos}, pioneered by Lasserre~\cite{lasserre2001global}. The approach of Lasserre has previously been applied to multiview geometry~\cite{kahlhenrion2007globallyoptimal-ijcv}, but without any tightness guarantees. In \cite{cifuentes-etal-2017}, SDP relaxations for quadratically constrained quadratic programs (QCQP) are analyzed.
Given that the SDP relaxation correctly solves a problem under noiseless observations (which is the case for the problems that we analyze), conditions are given which guarantee
strong duality
in the low noise regime.
The size of this neighborhood is however not explicitly given.
Further, a geometric interpretation of the relaxation is provided in \cite{cifuentes2020}. We base our framework on the mathematical results proved in \cite{blekherman-etal-2016} where a deep connection is established between, on one hand, algebraic varieties of minimal degree and on the other hand, the study of non-negativity and its relation with sums of squares (SOS).

\subsection{Contents of the paper}

In the next section,
we present our general problem
formulation.
In
Section~\ref{sec:applications},
applications are presented and formulated on the
standard form, of which the SDP relaxation
is given in Section~\ref{sec:sdprelaxation}.
In Sections~\ref{sec:duality}-\ref{sec:extremepoints}
we relate our problem to duality and analyze it using results from algebraic geometry. Our main result, a complete classification of SDP tightness for our example applications, is presented in Section~\ref{sec:tightness_applications}.

\section{Problem formulation}\label{sec:probform}
The class of problems that we are interested in analyzing are problems involving  rotations, parametrized either by
\begin{itemize}
    \item[(i)] $p\times p$ matrices belonging to the special orthogonal group, denoted $\SO{p}$, where $p=2$ or $p=3$ for planar rotations and 3D-rotations, respectively.
    \item[(ii)] $4$-vectors of unit length representing quaternions, denoted~$\Q$ for 3D-rotations.
\end{itemize}    
In addition we require that the objective function is quadratic in the variables of the chosen parametrization.

Let
$
R= [R_1, \ldots\,R_n]
$
where each $R_i \in \SO{p}$ with $p=2$ or $p=3$ and let $\vec{R}$ denote the column-wise stacked vector of the $p\times pn$ matrix $R$.
Now let $M$ be a real, symmetric $(p^2n+1)\times (p^2n+1)$ matrix, then we would like to solve the following non-convex optimization problem
\begin{equation}
\label{eq:mainoptSO}
\begin{array}{lll}
 &\min\limits_{R \in\SO{p}^n} & \begin{bmatrix}\vec{R}\\1\end{bmatrix}^T M \begin{bmatrix}\vec{R}\\1\end{bmatrix}
\end{array}.
\end{equation}
Alternatively we model 3D-rotations with unit quaternions,
$q= [q_1, \ldots\,q_n]$,
and consider
the similar formulation
\begin{equation}
 \label{eq:mainoptQ}
\begin{array}{lll}
 &\min\limits_{q \in \mathcal{Q}^n} & \begin{bmatrix}\vec{q}\\1\end{bmatrix}^T M \begin{bmatrix}\vec{q}\\1\end{bmatrix}
\end{array}.
\end{equation}
Note that not every problem with 3D-rotations may be straightforward to model on both of the formulations \eqref{eq:mainoptSO} and \eqref{eq:mainoptQ}, and furthermore, their residual errors have different interpretation and therefore the formulations are not equivalent. Also, the set of quaternions forms a double covering of the set of rotations meaning that $q$ and $-q$ represent the same rotations \cite{hartley2013}.

Both of these problem formulations can be put in the following standard form
\begin{equation}
 \label{eq:mainopt}
\begin{array}{lll}
 \min\limits_{r } & r^T M r \\
 {\mbox{subject to}} & r^T A_i r = 0, & i=1,\ldots,l\\
              & r^T e= 1
\end{array}.
\end{equation}
The
$l$ quadratic equations $r^T A_i r = 0$ enforce the rotational constraints
and $r^T e = 1$ with $e=[0,\ldots,0,1]^T$
forces the final element of $r$ to be one.
The rest of the paper will be devoted to
this standard form and we will analyze it in detail.

\section{Applications}\label{sec:applications}

There are several application problems that can be modelled and solved using the above formulation. Often, one would like to solve for one or several rigid transformations (a rotation and a translation). However, in many cases, one can directly eliminate the translation, and concentrate on the more difficult part of finding the rotations.

Next we give several examples of rotation problems appearing in the literature. 

\begin{example} \label{ex:registration} {\bf Registration} with point-to-point, point-to-line and point-to-plane correspondences can be written as in \eqref{eq:mainopt}.
The residuals are all of the form 
\begin{equation} \label{eq:registration}
\|P_i(R x_i + t - y_i)\|^2 = \|(x_i^T\otimes P_i)\vec{R} + P_i(t - y_i)\|^2.
\end{equation}
If point $x_i$ corresponds to point $y_i$ then set $P_i = I_3$. If $x_i$ is a measurement known to lie on a line then set $P_i = I - v_i v_i^T$, where $v_i$ is a unit directional vector and $y_i$ is a point on the line. 
Similarly, if $x_i$ lies on a plane then set $P_i = n_i^T$, where $n_i$ is a unit normal and $y_i$ is a point on the plane.

Minimizing over $t$ gives the closed-form solution
\newcommand{\tinysum}[1]{\scalebox{.8}{$\displaystyle\sum_{#1}$}}
\begin{equation}
t = -\Big(\tinysum{i} P_i^T P_i\Big)^{-1}\tinysum{i} P_i\left((x_i^T \otimes P_i)\vec{R} - P_i y_i\right),
\end{equation}
which is linear in $\vec{R}$. Inserting this into the objective function thus gives an expression which is quadratic in $\vec{R}$ and therefore can be reshaped into \eqref{eq:mainopt}.

In \cite{horn-etal-1987}, it is shown that the registration problem with point-to-point correspondences can be formulated as a quadratic optimization problem in the quaternion representation. We are not aware of any quadratic quaternion formulation for the point-to-line and point-to-plane cases.
\end{example}
\begin{example} \label{ex:resectioning} {\bf Resectioning} is the problem of recovering the position and orientation of a camera given 2D to 3D correspondences. Geometrically this can be done by aligning the viewing rays from the camera with the 3D points. This reduces the problem to a special case of point-to-line registration where all of the lines intersect in the camera center.
\end{example}

\begin{example} \label{ex:hand-eye} {\bf Hand-eye calibration} is the problem of determining the transformation between a sensor (often a camera) and a robot hand on which the sensor is mounted. Given rotation measurements $U_i$ and $V_i$, $i=1,\ldots,m$ relative to a global frame, for the sensor and the robot hand, respectively, the objective is to find the relative rotation $R$ between them by solving the following optimization problem:
\begin{equation} \label{eq:handeye}
\begin{array}{lll}
 &\min\limits_{R \in\SO{3}} & \sum\limits_{i=1}^m ||U_iR-RV_i||_F^2.
\end{array}
\end{equation}
$$
\begin{array}{lll}
    \| U_iR-RV_i \|_F^2 &=& \| (I\otimes U_i - V_i^T\otimes I)\vecemph{R} \|^2 \\
  &=&  \vecemph{R}^T M_i \vecemph{R},
\end{array}
$$
where $M_i = 2I- V_i\otimes U_i-V_i^T\otimes U_i^T$. Finally, set $M$ as
\begin{equation} \label{eq:M-handeye}
M = \begin{bmatrix} \sum_{i=1}^m M_i & 0 \\ 0 & 0 \end{bmatrix}.
\end{equation}
An alternative formulation using quaternions can also be employed.
If the unit quaternions $u,v \in \Q$ represent the rotations $U,V \in \SO{3}$ then the quaternion representing the composition $U V$ can be written $Q(u)v$ where 
\begin{equation}
    Q(u) = 
\mbox{\scriptsize$
    \left(
    \begin{matrix}
    u_0 & -u_1 & -u_2 & -u_3\\
u_1 & u_0 &  -u_3 & u_2\\
u_2 & u_3 &  u_0 & -u_1\\
u_3 & -u_2 & u_1 &  u_0 
    \end{matrix}
    \right).
$}
\end{equation}
Therefore the optimization problem 
\begin{equation}
\begin{array}{lll}
 &\min\limits_{q \in \Q} & \sum\limits_{i=1}^m ||Q(u_i)q-Q(q)v_i||^2
\end{array}
\end{equation}
can also be used and turned into the standard form in order to solve hand-eye calibration. Note however that due to the double covering the signs of $u_i$ and $v_i$ have to be selected consistently in order for $q$ to give a low objective value.
\end{example}

\begin{example} {\bf Rotation averaging} aims to determine a set of absolute orientations $R_i$, $i=1,...,n$ from a number of measured relative rotations $R_{ij}$ by minimizing
\begin{equation}
\sum_{i \not= j}  \|R_i R_{ij}-R_j\|_F^2.
\end{equation}
Since $\|R_i\|_F^2 = 3$ the problem is (ignoring constants) equivalent to minimizing
\begin{equation} \label{eq:rotavobj_and_M0}
\begin{array}{cc}
    \begin{array}{c}
        -{\displaystyle\sum_{i \not = j}} \langle R_iR_{ij},R_j\rangle \\[1.5em]
        = \tr{R M_0 R^T},
    \end{array}
    & , \quad
    M_0 = -
    \mbox{\scriptsize$
    \begin{bmatrix}
    0 & R_{12} & \hdots & R_{1n} \\
    R^T_{12} & 0 & \hdots & R_{2n} \\
    \vdots & \vdots & \ddots & \vdots \\
    R^T_{1n} & R^T_{2n} &  \hdots & 0 \\
    \end{bmatrix}.
    $}
\end{array}
\end{equation}

Letting $M = \blkdiag{M_0 \otimes I_3, 0}$ now gives an optimization problem of the form \eqref{eq:mainopt}. $I_3$ is a $3 \times 3$ identity matrix and the $\blkdiag{\cdot}$ operation constructs a block-diagonal matrix.

Similar to the hand-eye calibration problem, rotation averaging can be formulated with quaternions \cite{govindu-cvpr-2001} using the objective function
$\sum_{i\neq j} \|Q(r_i)r_{ij} - r_j\|^2$,
which after simplifications yields an expression similar to \eqref{eq:rotavobj_and_M0} and hence can be put in the standard form \eqref{eq:mainopt}.
\end{example}

\begin{example}{\bf Point set averaging} is the problem of registering a number of point sets, measured in different coordinate systems, to an unknown average model.
If $X_i$, $i=1,...,n$ are $3 \times n$ matrices containing measurements of corresponding $3D$ points, we want to find a $3 \times n$ matrix $Y$, rotations $R_i$ and translations $t_i$ such that 
\begin{equation}
\sum_i \|R_i X_i + t_i\ones^T - Y\|_F^2,
\end{equation}
where $\ones$ is a column vector with all ones, is minimized.
Since the variables $Y$ and $t_i$, $i=1,...,n$ are unconstrained, they can be solved for as a function of the rotations $R_i$, $i=1,...,n$. Assuming that the centroid of the points in $X_i$ is the origin for all $i$, back-substitution allows us to write the problem solely in terms of the rotations as
\begin{equation}
\min_{R \in \SO{3}^n} \tr{R M_0 R^T},
\end{equation}
where
\begin{equation}
M_0 =
\mbox{\scriptsize$
-\left[\begin{array}{ccccc}
0 & X_1 X_2^T & X_1 X_3^T & \hdots & X_1 X_n^T\\
X_2 X_1^T & 0 & X_2X_3^T & \hdots & X_2 X_n^T\\
\vdots & \vdots & \vdots & \ddots & \vdots \\
X_n X_1^T & X_n X_2^T & X_n X_3^T & \hdots & 0
\end{array}\right].
$}
\end{equation}
Letting $M = \blkdiag{M_0 \otimes I_3, 0}$ now gives an optimization problem of the form \eqref{eq:mainopt}.
\end{example}
In the above examples we only considered the case of 3D-rotations. Note however that all of these problems also have meaningful versions in the plane for which parametrization using $\SO{2}$ yields the same type of problem.

\section{SDP relaxation}\label{sec:sdprelaxation}

Let us now derive the standard convex SDP relaxation for our standard form~\eqref{eq:mainopt}. Consider the objective function, which can be rewritten using trace notation as
$$
\begin{array}{lllll}
r^T M r &=& \tr{ Mrr^T} &=& \tr{MX},
\end{array}
$$
where $X=rr^T$. %
Note that $\rank{X}=1$ and %
$X \succeq 0$.

For $\SO{3}$, the condition that $R_i$ belongs to the special orthogonal group can be expressed by quadratic constraints in the entries of $R_i$, for instance $R_i^TR_i-I=0$. Similarly, that the cross product of the first and second rows should equal the third, which ensures $\det{R_i}=1$, is also a quadratic constraint. Consequently, the same constraints can be expressed by linear equations in the entries of $X$ in the form $\tr{A_iX}=0$. It can be checked that there are $20$ linearly independent such constraints for each $R_i$.

The corresponding program for $\SO{2}$ is similar to that of $\SO{3}$ but it only requires one constraint per rotation. We represent a rotation by
$ R_i = \left(
\begin{matrix}
c_i & -s_i \\ s_i & c_i
\end{matrix}
\right),
$
where $c_i^2+s_i^2 = 1$. Hence, for $n$ rotations, only $2n$ variables are needed in the vector $r$ and the unit length constraint becomes linear in the entries of $X$. %
Similarly for quaternions $\Q$, the unit length constraint for each $q_i$ can be written
as a linear constraint. %

If we ignore the non-convex constraint that $\rank{X}=1$, then we get a semidefinite program problem over $X$: The objective function is linear in $X$ subject to linear equality constraints and a positive semidefinite constraint, $X\succeq 0$.
This leads to the following convex relaxation: %
\begin{equation} \label{eq:relaxation}
\begin{array}{lll}
 &\min\limits_{X \succeq 0 } &
 \tr{MX} \\
 & & \tr{A_iX} = 0, \quad i=1,\ldots,l \\
 & &     \tr{ee^TX} = 1
\end{array}.
\end{equation}
Note that as we have relaxed (ignored) $\rank{X}=1$, the optimal value will be a lower bound on the original non-convex problem (\ref{eq:mainopt}). Further, if the optimal solution $X^*$ has rank one, then we say that the relaxation is tight and the globally optimal solution is obtained.

\section{Duality and sums of squares}\label{sec:duality}

Consider again the objective function in (\ref{eq:mainopt}) and the Langrangian dual problem of \eqref{eq:relaxation}
\begin{equation}
\begin{array}{lll}
 &\max\limits_{\gamma,\lambda_1,\ldots,\lambda_l } &
 \gamma \\
 & & M-\sum_i \lambda_i A_i - \gamma ee^T \succeq 0,
\end{array}
\label{eq:dual}
\end{equation}
where $(\gamma,\lambda)$ are the dual variables. By construction this problem gives the same objective value as \eqref{eq:relaxation} and therefore a lower bound on the original \eqref{eq:mainopt}. We are interested in knowing when this lower bound is tight.

Let $I$ be the ideal of the polynomials defining the constraint set, that is, a polynomial $p $ is in $I$ when
\begin{equation}
p(r) = v(r)(r^Te e^T r-1) + \sum_i v_i(r)r^T A_i r,
\end{equation}
where $v$ and $v_i$ are any polynomials. The variety $V(I)$ consists of the feasible points $\{r \, | \, p(r) = 0\ \forall p\in I\}$. Let $R_2$ denote the set of quadratic polynomials
modulo $I$, that is,
two polynomials $f, g \in R_2$ are considered equal if $f-g \in I$.

The question of tightness between the original problem~\eqref{eq:mainopt} and the relaxation~\eqref{eq:relaxation} and its dual \eqref{eq:dual} is related to the two convex, closed cones
\begin{equation}
    \label{eq:nn-cone}
 P := \left\{ f \in R_2 \, | \, f(r) \ge 0 \mbox{ for all } r \in V(I) \,\right\},
\end{equation}
and 
\begin{equation}
\begin{array}{ll}
    \label{eq:sos-cone}
\Sigma := \{ f \in R_2 \, | \,& \mbox{there exist vectors } a_1,\ldots,a_k \\ &\mbox{such that } f(r) = \sum_{i=1}^k (a_i^T r)^2 \, \}.
\end{array}
\end{equation}
Note that the cones are defined to be dependent on the constraint set of~\eqref{eq:mainopt}, and not on the actual form of the objective function $r^TMr$. As any quadratic polynomial $f$ in $\Sigma$ is a sum of linear squares on the feasible set $V(I)$ and hence non-negative, it follows that $\Sigma \subseteq P$.

Consider again our original problem in (\ref{eq:mainopt}), written as
$$\eta^* = \min_{r\in V(I)} r^T M r.$$
It follows that $r^T(M-\eta^*ee^T)r \in P$. If $\gamma^*$ is the optimal value of \eqref{eq:dual} with dual variables $\lambda^*$, then the matrix $M-\sum_i \lambda^*_i A_i - \gamma^* ee^T$ is positive semidefinite and we can factor it into 
a sum of rank-1 matrices $\sum_j a_j a_j^T$. Therefore,
$$
\sum_j (a_j^Tr)^2 = r^T (M-\gamma^*ee^T) r - r^T\left(\sum_i \lambda^*_i A_i\right)r .
$$
Now, $r^T\left(\sum_i \lambda^*_i A_i\right)r$ lies in $I$ and we can conclude that the quadratic polynomial $r^T(M - \gamma^*ee^T)r$ belongs to $\Sigma$ when $(\gamma^*,\lambda^*)$ is the solution to \eqref{eq:dual}.

In view of the above discussion it is clear that the convex formulations \eqref{eq:relaxation} and \eqref{eq:dual} can only give the same objective value as \eqref{eq:mainopt} when $r^T M r - \eta^*$ is a sum of squares, where $\eta^*$ is the optimal value of \eqref{eq:mainopt}.
The question we are interested in answering is hence {\em when is it possible to find an SOS for this non-negative quadratic form}?
If the cones are not equal, that is, $\Sigma \subsetneq P$, then there may exist objective functions for which the relaxation is not tight. We shall investigate this further in a constructive manner. First, we need some more tools from algebraic geometry. See also the book \cite{convalg-book-2012} for a general introduction.

\section{The varieties of rotations}\label{sec:variety}

An algebraic variety $V$ is the set of solutions of a system of polynomial equations over the reals. In this paper we analyse three varieties that are commonly used in computer vision applications: $\SO{2}^n$, $\SO{3}^n$ and $\Q^n$. These varieties can be defined by a system of polynomial equations in the entries of $2\times 2$, $3\times 3$ matrices and 4-vectors, respectively (cf.~Section~\ref{sec:sdprelaxation}). The dimensions and co-dimensions of these varieties are well-known, and $\dim{\SO{2}} = 1$,  $\codim{\SO{2}} = 1$,  $\dim{\SO{3}} = 3$,  $\codim{\SO{3}} = 6$, $\dim{\Q} = 3$, and $\codim{\Q} = 1$. The degree of $V$ is by definition the number of intersection points of the variety with $\dim{V}$ general hyperplanes, and we have that $\deg{\SO{2}} = 2$, $\deg{\SO{3}} = 8$ (see \cite{brandt-etal-2017} for a derivation) and $\deg{\Q} = 2$.

For $n$ copies of $V$, it is straightforward to show that the variety of $V^n$ has dimension $n\dim{V}$, co-dimension $n \codim{V}$, and degree $(\deg{V})^n$. For instance, for the case of $\SO{3}^n$, we have that $\dim{\SO{3}^n} = 3n$, $\codim{\SO{3}^n} = 6n$, and $\deg{\SO{3}^n} = 8^n$. 

For any variety $V$, $\deg{V} \geq \codim{V}+1$.
A variety is called {\em minimal} if it is non-degenerate (that is, not contained in a hyperplane) and $\deg{V} = \codim{V}+1$. Similarly, it is called {\em almost minimal} when $\deg{V} = \codim{V}+2$.
Considering the degrees and co-dimensions of the varieties previously listed, Table~\ref{tab:minitable} summarizes their characterization as minimal, almost minimal, and not minimal, for the cases $n = 1$, $n = 2$, and $n > 2$. 

\begin{table}[t!]
    \centering
    \begin{tabular}{|c|c|c|c|}
    \hline
    & $n=1$ & $n=2$ & $n>2$ \\
    \hline
    $\SO{3}^n$ & \amcell & \nmcell & \nmcell \\
    \hline
    $\SO{2}^n$ & \mcell & \amcell & \nmcell \\
    \hline
    $\Q^n$ & \mcell & \amcell & \nmcell \\
    \hline
    \end{tabular}
    \caption{Characterization of $V=\SO{3}^n$, $V=\SO{2}^n$ and $V=\Q^n$, in terms of their degree. If $\deg{V}=\codim{V}+1$, $V$ is said to be \emph{minimal}, and if $\deg{V}=\codim{V}+2$, $V$ is \emph{almost minimal}. Otherwise $\deg{V} \geq \codim{V}+3$, and $V$ is neither \emph{minimal} nor \emph{almost minimal}.}
    \label{tab:minitable}
\end{table}

\section{The extreme points of the SDP relaxation}\label{sec:extremepoints}
In this section, we investigate further the convex cone of non-negative polynomials and that of SOS polynomials over the rotational varieties, $\SO{3}$, $\SO{2}$ and $\Q$. The goal is to find out when the SDP relaxation is tight and to characterize all possible extreme points for the relaxation.

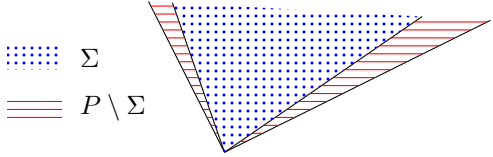
\begin{figure}[t]
    \centering
      \begin{tikzpicture}
  [sos_pattern/.style ={pattern color=blue, pattern=dots},
   nonsos_pattern/.style ={pattern color=red, pattern=horizontal lines}]

  \coordinate (origin) at (0,0);
  \coordinate (left_outer) at (-1,2);
  \coordinate (right_outer) at (3.5,1.75);
  \coordinate (left_inner) at ($(left_outer)!0.07!(right_outer)$);
  \coordinate (right_inner) at ($(left_outer)!0.80!(right_outer)$);
  \coordinate (left_outer) at (-1,2);

  \coordinate (sos_pattern_coord) at (-2.5,1.25); %

  \draw (origin) -- (left_outer);
  \draw (origin) -- (left_inner);
  \draw (origin) -- (right_inner);
  \draw (origin) -- (right_outer);

  \fill[nonsos_pattern] (origin) -- (left_inner) -- (left_outer) -- cycle;
  \fill[sos_pattern] (origin) -- (left_inner) -- (right_inner) -- cycle;
  \fill[nonsos_pattern] (origin) -- (right_inner) -- (right_outer) -- cycle;

  \node[rectangle, sos_pattern, align=left, minimum width=0.7cm, minimum height=0.3cm, ultra thick] (sos_pattern) at (sos_pattern_coord) {};
  \node[align=left] (sos_text)[right=0.1cm of sos_pattern] {$\Sigma$};

  \node[rectangle, nonsos_pattern, align=left, minimum width=0.7cm, minimum height=0.3cm, ultra thick] (nonsos_pattern)[below=0.3cm of sos_pattern] {};
  \node[align=left] (nonsos_text)[right=0.1cm of nonsos_pattern] {$P \setminus \Sigma$};

  \end{tikzpicture}
    \caption{Illustration of the closed convex cones $\Sigma$ and $P$ defined in (\ref{eq:nn-cone}) and (\ref{eq:sos-cone}). If the variety $V(I)$ is minimal then $\Sigma=P$, otherwise $\Sigma \subsetneq P$.}
    \label{fig:cone_fig}
\end{figure}

The following general result is proved in \cite{blekherman-etal-2016}.
\begin{lemma} \label{lem:minimal} (Blekherman et\ al.\ \cite{blekherman-etal-2016})
Let $V$ be a real irreducible non-degenerate variety such that its subset of real points is Zariski dense.
Every real quadratic form that is non-negative on $V$ is a sum of squares of linear forms if and only if $V$ is a variety of minimal degree.
\end{lemma}
An illustration of the result is given in Figure~\ref{fig:cone_fig}. We will now apply it to our varieties of interest\footnote{All of the varieties that we study are real, irreducible, non-degenerate and their corresponding subsets of real points are Zariski dense.}.

\subsection{Minimal varieties}
Among the varieties in Table~\ref{tab:minitable}, we know that only $\SO{2}$ and $\Q$ are minimal. In the remaining cases, the convex cones $P$ and $\Sigma$ defined in (\ref{eq:nn-cone}) and (\ref{eq:sos-cone}) are therefore strictly different, i.e. $\Sigma \subsetneq P$. The proof of Lemma~\ref{lem:minimal} is constructive and it allows us to generate objective functions that are non-negative, but not sums of squares, and thereby the SDP relaxations will not be tight for these optimization problems. However, our example applications have objective functions of a special form and it remains to see if there are such objectives which are not sums of squares. We will return to this question in Section~\ref{sec:tightness_applications}.
We can conclude that any hand-eye calibration problem defined over $\SO{2}$ or $\Q$ will always have tight SDP relaxations.

\subsection{Almost minimal varieties}
In the case when $V$ is \emph{almost minimal}, that is, when $V$ is either $\SO{3}$, $\SO{2} \times \SO{2}$ or $\Q \times \Q$ (Table~\ref{tab:minitable}), we will still have $\Sigma \subsetneq P$, but the gap between the cones will be smaller.
Furthermore, for problems in $P \setminus \Sigma$, the extreme points of the corresponding SDP relaxation can be characterized based on the theory in \cite{blekherman-etal-2016}. 
An immediate reformulation of Proposition~3.5 for our purposes gives the following corollary.
\begin{corollary} \label{th:extremepoints} Assume that the variety $V$ is almost minimal and 
arithmetically Cohen-Macaulay. Then, the extreme points $X^*$ of the SDP relaxation in (\ref{eq:relaxation}) either have $\rankemph{X^*}=1$ or $\rankemph{X^*}=\codim{V}$.
\end{corollary}

All of the varieties we study are smooth, and therefore arithmetically Cohen-Macaulay. Furthermore recall from Section~\ref{sec:variety} that $\codim{\SO{3}} = 6$, $\codim{\SO{2} \times \SO{2}} = 2$ and $\codim{\Q \times \Q} = 2$.
In the $\SO{3}$ case, if the computed optimal solution $X^*$ of the relaxation has not rank 1 nor 6, but say for instance rank 2, then $X^*$
can be decomposed into two rank-1 matrices, $X^*=\lambda X_1^*+(1-\lambda)X_2^*$
for some $\lambda \in [ 0,1 ]$ where $X_1^*$ and $X_2^*$ are optimal solutions and extreme points.

If $\rank{X^*}=1$ then the corresponding objective function $r^TMr-\eta^*$ (where $\eta^*$ is the optimal objective value) is a sum of squares, and as shown previously can be retrieved by solving the SDP. If $\rank{X^*}>1$ and $X^*$ cannot be decomposed into rank-1 extreme points, then the corresponding objective function $r^TMr-\eta^*$ is not a sum of squares. For almost minimal arithmetically Cohen-Macaulay varieties, such extreme points $X^*$ must be of $\rank{X^*}=\codim{V}$ and there are no other possibilities.

To summarize, if when minimizing a given problem over an \emph{almost minimal} variety $V$ we obtain the extreme point
$X^*$ which has $\rank{X^*}=1$, then we have indeed computed the globally optimal solution, but if it turns out that $\rank{X^*}=\codim{V}$, then the relaxation is not tight, and we do not even have a feasible solution to the original problem, just a lower bound on the optimal value.

\subsection{Prevalence of non-tight problem instances}\label{sec:examples}
In Figure~\ref{fig:empirical1}, we presented the results of two sets of synthetic experiments, illustrating the significance of \emph{almost minimal} varieties.

In the first set of experiments, the domain is the \emph{almost minimal} variety $V=\SO{3}$ and the entries of the objective function, encoded by the $10\times10$ symmetric matrix $M$ in (\ref{eq:mainopt}), were randomly drawn from a uniform distribution from $[\,\text{-}1,1\,]$. In all $1000$ examples, we obtained a rank-1, globally optimal solution for the SDP relaxation, even though the variety is not minimal.
This shows that the rank-6 extreme points predicted by Corollary~\ref{th:extremepoints} are rare in practice among the random objective functions considered. It is however possible to produce such non-tight examples and we shall return to this question later.

In the second set of experiments, the optimization took place over $V=\SO{3} \times \SO{3}$, which is not \emph{almost minimal}. The entries of the $19 \times 19$ symmetric matrix $M$ were generated in the same way via a uniform distribution. In this case, the relaxation works poorer and various ranks
are obtained for its solutions.

\paragraph{Remark. }
For neither \emph{minimal} nor \emph{almost minimal} varieties, the non-negative cone $P$ becomes significantly larger than the SOS cone $\Sigma$.
Non-tight SDP relaxations will be more prevalent, and various ranks will be observed for the
solutions to
these non-tight relaxations.
A rank-1 solution will however always provide a solution to the primal problem.

\section{Tightness of our example applications}\label{sec:tightness_applications}

\newcommand{\gcell}[1]{\cellcolor{green!25} #1}
\newcommand{\ycell}[1]{\cellcolor{yellow!25} #1}
\newcommand{\rcell}[1]{\cellcolor{red!25} #1}
\newcommand{\nottight}{Not always tight}
\newcommand{\lownoise}[1]{$\sigma$:~#1}
\newcommand{\lnsep}{\raisebox{-.2em}{\large /}}

 \begin{table*}[t!]
     \centering
     \begin{tabular}{|c|c|c|c|c|}
     \hline
     & Registration $\&$ resectioning & Hand-eye calibration & Rotation averaging & Point set averaging  \\
     \hline
     $\SO{3}$
         & \ntcell
         & \ntcell
         & \lncell~\cite{rosen-etal-2018,cifuentes-etal-2017,eriksson-etal-pami-2019}
         & \lncell~\cite{Chaudhury-etal-siam-2015,iglesias-etal-cvpr-2020}
         \\
     \hline
     $\SO{2}$
         & \tcell
         & \tcell
         & \lncell~\cite{zhong-boumal-siam-2018} %
         & \lncell~\cite{Chaudhury-etal-siam-2015}
         \\
     \hline
     $\Q$
         & \tcell
         & \tcell
         & \lncell~\cite{cifuentes-etal-2017}
         & Not applicable
         \\
     \hline
     \end{tabular}
     \caption{
     Tightness of SDP relaxations for various applications
     and
     parametrizations.
     Colors follow
     Table~\ref{tab:minitable}, illustrating whether the domain is \emph{minimal}, \emph{almost minimal}, or neither.
     The main new results are for the
     \emph{almost minimal} cases,
     for which we have generated rare non-tight counterexamples.
     For the {\em low noise} cases,
     tightness
     can only be guaranteed
     in the low noise regime.
     We conclude
     that only the problem classes over
     \emph{minimal} varieties
     come with tightness guarantees.
     }
     \label{tab:minitable2}
 \end{table*}

The theoretical results in the previous section apply to general quadratic objective functions. For actual applications, the objective functions will be structured. For instance, consider the hand-eye calibration problem in Example~\ref{ex:hand-eye}. There are only purely quadratic terms of the rotation variables in the objective, and no linear ones. Hence, the last row and the last column of the matrix $M$ will be zero. In this section, we analyze structured objective functions corresponding to different problem classes. We also relate our new results to previous ones in the literature.

In Table~\ref{tab:minitable2} we present a complete classification of SDP tightness for our example applications.
In accordance with Table~\ref{tab:minitable}, applications for the \emph{minimal} varieties $\SO{2}$ and $\Q$ are always tight --
this is a known result, as there is a single quadratic constraint (see e.g. Boyd and Vandenberghe~\cite{boyd-vandenberghe-book-2004}).
For the \emph{almost minimal} varieties, we generate rare non-tight problem instances, and for the \emph{non-minimal} cases we conclude that tightness can only be guaranteed in the low noise regime,
supported by previous works and empirically demonstrated by us.

\paragraph{Noise-free case. } All of our considered example applications have objective functions
of the form $r^TMr = r^TU^TUr = \| Ur \|^2$ for some matrix $U$. If the optimal value $\eta^* = \min_{r\in V(I)} r^T M r$ is equal to zero (which corresponds to the noise-free case) then $r^T(M-\eta^*ee^T)r = r^TMr \in P$, where $P$ is the cone of non-negative quadratic forms in (\ref{eq:nn-cone}). Further, since $M$ has non-negative
eigenvalues, $M\succeq 0$ and we
can factor it into a sum of rank-1 matrices $M = \sum_j a_j a_j^T$.
It follows that
$r^TMr = \sum_j (a_j^Tr)^2$ is
SOS.
This is a well-known result and
it has further been studied
in \cite{cifuentes-etal-2017} where it is shown that for low noise levels ($\eta^*$ close to zero), the non-negative polynomial $r^T M r - \eta^*$ is a sum of squares as well.

\subsection{Registration and resectioning}
The formulations of these two applications over the domain $\SO{3}$ are given in Examples~\ref{ex:registration} and~\ref{ex:resectioning}, respectively. As the variety $\SO{3}$ is almost minimal (Table~\ref{tab:minitable}), one may wonder if there are actual problem instances that lead to non-tight relaxations and extreme points of rank-6 (Corollary~\ref{th:extremepoints})? In \cite{blekherman-etal-2016}, there is a procedure for finding polynomials that are non-negative, but not sums of squares. However, this will in
general not result in objective functions originating from registration or resectioning problems. The objective function for this type of problem is of the form $r^TMr = r^TU^TUr = \| Ur \|^2$ where $M=U^TU$ with $U$ of size $m \times 10$ and $m$ is the number of correspondences of type point-to-point, point-to-line or point-to-plane. That in particular means that $M$ has non-negative eigenvalues and $M\succeq 0$, and there are some additional requirements as well.

In Section~\ref{sec:generating_nontight_ls_problems} in the appendix, we show how to modify the procedure of Blekherman et\ al.~\cite{blekherman-etal-2016} in order to achieve such objective functions. For every non-tight problem instance generated with Procedure~\ref{proc:P_not_Sigma} described in Section~\ref{sec:generating_nontight_ls_problems}, we get a rank-6 solution $X^*$ as predicted by Corollary~\ref{th:extremepoints} and consequently no feasible solution is obtained. Hence, there exist indeed problem instances that are non-tight, but they are rare in practice. See also the first column of Table~\ref{tab:minitable2}.

\paragraph{Relation to the empirical results of Briales and Gonzalez-Jimenez \cite{briales-cvpr-2017}. } Extensive experiments using the SDP relaxation in \eqref{eq:relaxation}
for registration over $\SO{3}$ are performed in \cite{briales-cvpr-2017}, but not a single instance with a non-tight relaxation among their real or synthetic experiments is found. This is consistent with our experiments, presented in Figure~\ref{fig:empirical1}(a), where we have done an empirical analysis of SDP tightness over $\SO{3}$ for quadratic objective functions with random entries.
The counterexamples are indeed rare in practice for this almost minimal variety.

\subsection{Hand-eye calibration}
As previously mentioned, the objective function for hand-eye calibration contains only purely quadratic terms of the rotation variables and no linear ones. Hence, the last row and the last column of the $10\times10$ symmetric matrix $M$ will be zero. We tested the same procedure as for registration (see Procedure~\ref{proc:P_not_Sigma} in Section~\ref{sec:generating_nontight_ls_problems} in the appendix) in order to generate non-tight counterexamples with the structure of a hand-eye calibration objective. We succeeded in obtaining problem instances for averaging 8 rotation matrices that yielded objective functions with non-tight relaxations, see the second column of Table~\ref{tab:minitable2} for a summary.
Again, all of these optimization problems attain their minima at rank-6 extreme points, in accordance with Corollary~\ref{th:extremepoints}.

\subsection{Rotation averaging}
In \cite{eriksson-etal-pami-2019}, Eriksson et\ al.\ proved that the SDP relaxation for problems involving three rotations with $\SO{3}$-parametrization is always tight.
This result trivially extends to $\SO{2}$.
Further, for instances with more than three cameras, it is shown that in the low noise regime the SDP relaxation is tight.
Low noise results applicable to $\SO{3}$ as well as $\SO{2}$ have also been presented by Rosen et\ al.~\cite{rosen-etal-2018}, although $\SO{2}$ is parametrized by all matrix elements in their case.
Fredriksson and Olsson \cite{fredriksson2012simultaneous} parametrize the rotation averaging problem with quaternions $\Q$ and in all the reported experiments, the SDP relaxation was always numerically found to be tight. For $\SO{2}$,
Zhong and Boumal~\cite{zhong-boumal-siam-2018} proved the existence of an upper bound on the noise level for which the SDP relaxation is tight, however no explicit estimates were
given\footnote{\label{fn:complex-repr-equiv}Their $\mathbb{C}$ parametrization is equivalent to our representation.}.

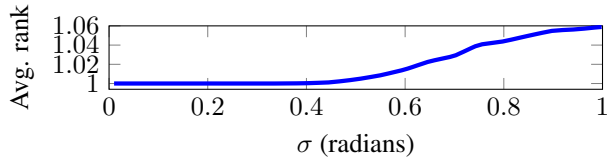
\begin{figure}
    \centering
    \begin{tikzpicture}
\begin{axis}[
  width=\linewidth,
  height=.3\linewidth,
  ytick={1,1.02,1.04,1.06},
  xlabel={$\sigma$ (radians)},
  ylabel={Avg. rank},
  xmin=0,
  xmax=1,
  ymax=1.06,
]
\addplot[mark=none, blue, ultra thick] table [x=x, y=y, col sep=comma] {fig3.txt};
\end{axis}
\end{tikzpicture}
    \caption{Average rank for instances of the rotation averaging problem over $\SO{2}^4$, with varying noise levels.}
    \label{fig:rotation_averaging_rank}
\end{figure}

Here we present results for the case of four rotations in $\SO{2}$
(a three rotation problem is always tight \cite{eriksson-etal-pami-2019}).
Figure~\ref{fig:rotation_averaging_rank} shows the average rank of
the
computed SDP solution $X^*$. The $M_0$ matrix in (\ref{eq:rotavobj_and_M0}) was generated by
sampling
the relative rotation angles
from $\mathcal{N}(0, \sigma^2)$, $\sigma \in [0, 1]$ radians.
For each noise level $\sigma$, we ran the problem $10,000$ times and plotted the average obtained rank of the lifted variables $X^*$. The observed ranks were $1$ or $2$.
Similar to our results, Fan et\ al.~\cite{fan2020} find instances of the 2D SLAM problem with non-tight relaxations\footnoteref{fn:complex-repr-equiv}, and Carlone et\ al.~\cite{carlone-etal-iros-2015} present analogous results for
3D pose-graph optimization (PGO).

\paragraph{Relation to Mangelson et\ al.\ \cite{mangelson-etal-icra-2019}.} 
The planar pose-graph problem with
\begin{equation}\label{eq:pose_graph_objective_mangelson}
\sum_{i \neq j} \|R_i R_{ij} -R_j\|_F^2 + \tau \|t_j - t_i - R_i t_{ij}\|^2
\end{equation}
is studied in \cite{mangelson-etal-icra-2019}. 
Here, additional relative translation estimates $t_{ij}$ are present,
but $\tau = 0$ reduces the problem to rotation averaging.
A `proof' of strong duality for the Sparse-BSOS relaxation \cite{lasserre-etal-2015,weisser-etal-2016} is presented.
This would imply that rotation-averaging can be solved exactly in polynomial time but that the SDP relaxation~\eqref{eq:relaxation} still gives a duality gap.
While such a weakness is entirely plausible we note that the presented proof in \cite{mangelson-etal-icra-2019} is in fact flawed as the domain,
$\SO{2}^n$ using unit norm constraints on the diagonals of the rotation matrices, is incorrectly claimed to be SOS-convex (see \cite{convalg-book-2012} for a definition).
It is not even a convex domain.

While the lack of a proof does not exclude the possibility that Sparse-BSOS is exact, our counterexamples in Figure~\ref{fig:rotation_averaging_rank} show that this is only possible if Sparse-BSOS is stronger than the SDP relaxation~\eqref{eq:relaxation}.
A detailed comparison of these two formulations would reveal if this is the case.
Such an in depth analysis is however beyond the scope of this paper.

\subsection{Point set averaging}
\begin{figure}[h]
    \centering
    \begin{tikzpicture}
\begin{axis}[
  width=\linewidth,
  height=.35\linewidth,
  ytick={1,3,5,7,9},
  xlabel={$\sigma$ (rel. to signal)},
  ylabel={Avg. rank},
  xmin=0,
  xmax=2,
  ymax=9,
]
\addplot[mark=none, blue, ultra thick] table [x=x, y=y, col sep=comma] {fig4.txt};
\end{axis}
\end{tikzpicture}
    \caption{Average rank for instances of the point set averaging problem over $\SO{3}^4$, with varying noise levels.}
    \label{fig:point_averaging_rank}
\end{figure}
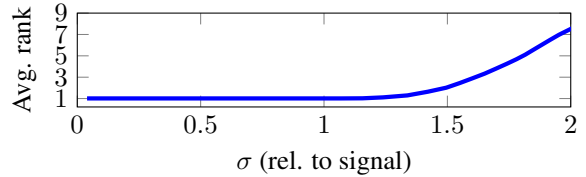

In previous work by Chaudhury et\ al.\ and Iglesias et\ al.\ \cite{Chaudhury-etal-siam-2015,iglesias-etal-cvpr-2020}, it has been shown that SDP relaxations are tight in the low noise regime for registering multiple
point clouds, while in the high noise regime non-tight instances arise.
Here we reproduce similar results, registering
an artificial point set over four frames.
$100$
points are
sampled from $\mathcal{N}(0, 1)$, after which
one direction is squeezed with a factor $1/100$,
causing
higher prevalence of non-tight instances.
Gaussian noise, sampled from $\mathcal{N}(0, \sigma^{2})$, was added to
each point.
Figure~\ref{fig:point_averaging_rank} shows the average rank over $10.000$ problem instances of the computed SDP solution $X^*$, for each noise level $\sigma$.

\section{Conclusions}

We have presented a framework for analyzing the power of
SDP
relaxations for optimization over rotational constraints. The key to our analysis has been to investigate the two convex cones of non-negative and sum-of-squares polynomials, and to establish a connection between them and the tightness of an SDP relaxation. We have shown that certain parametrizations lead to tight SDP relaxations and others do not. For our applications which have structured objective functions, we have generated non-tight counterexamples to settle the question of whether the relaxation
is always tight or not.

An interesting avenue for future research is to develop algorithms that can recover a good solution from a non-tight relaxation for practitioners of SDP relaxations. This was recently done for the rotation averaging problem \cite{dellaert-eccv-2020}.
Another interesting direction is to explore the existence of noise bounds for which the registration and hand-eye
calibration problems over $\SO{3}$ are guaranteed to be
tight.

\section*{Acknowledgements}
This work was partially supported by the Wallenberg AI, Autonomous Systems and Software Program (WASP), %
the Swedish Foundation for Strategic Research
and the Swedish Research Council (grant nr. 2018-05375).
Special thanks for the insightful and invaluable comments of our reviewers.

\clearpage
\appendix

\section{Generating non-tight least-squares problems}\label{sec:generating_nontight_ls_problems}

Using Procedure~\ref{proc:P_not_Sigma}, one can generate non-negative polynomials that are not sums of squares, and consequently such objective functions minimized over the variety will result in non-tight relaxations. The procedure is adapted from Procedure~3.3 in \cite{blekherman-etal-2016}
(for general non-minimal varieties),
and presented here specifically for the $\SO{3}$-case. We will generate explicit counterexamples for the hand-eye calibration problem and the registration problem that do not have tight SDP relaxations.

\begin{procedure}
\caption{Non-negative polynomials that are not sums of squares, $p \in P \setminus \Sigma$.}\label{proc:P_not_Sigma}
This procedure gives an objective function $r^TMr$ which is not an SOS and for which the optimal solution is a rank-6 extreme point. Recall that $r$ is a $10$-vector consisting of $9$ variables and the constant term~$1$ and $M$ is a $10\times10$ symmetric matrix.
\begin{enumerate}
    \item Choose general linear forms $h_i^Tr$, $i=1,\ldots,\dim{V}$, where $\dim{V}=3$ and compute the $\deg{V}=8$ intersection points with the variety $\SO{3}$, where all the intersection points should be real. Fix $\codim{V}=6$ of the points and choose an additional linear form $h_0^Tr$ that vanishes on these 6 points.
    \item Choose a quadratic form $r^TM_0r$ that (i) vanishes to order at least two at each of the 6 selected intersection points and (ii) does not belong to the subspace generated by the forms $(h_i^Tr)(h_j^Tr)$ for $i,j \in \{0,1,2,3\}$.
    \item For every sufficiently small $\delta$, the quadratic form 
    $$
    r^TMr = \delta r^TM_0r + (h_0^Tr)^2+ (h_1^Tr)^2+ (h_2^Tr)^2+ (h_3^Tr)^2
    $$
    is non-negative on $\SO{3}$ but not a sum of squares.
\end{enumerate}
\end{procedure}

There are a few modifications of the procedure that are required so that the objective functions do originate from a specific example application. For least-squares problems (such as hand-eye and registration), we know that $M \succeq 0$. This is not automatically fulfilled via Procedure~\ref{proc:P_not_Sigma}, but we can change Step~2 to account for this. More specifically, one can choose a quadratic form $M_0$ which fulfills the two prescribed conditions (i) and (ii), and at the same time maximizes the minimum eigenvalue of $M_0$. This can be cast as a convex optimization problem. If an $M_0 \succeq 0$ is found, then $M \succeq 0$ will follow. Further, for hand-eye calibration and registration, there are linear relationships between the elements of $M$ that should be satisfied for any problem instance, which can be accounted for when maximizing the minimum eigenvalue of $M_0$.%

To show that such an $M$ corresponds to an actual problem instance, we look for, in the case of hand-eye calibration, pairs of rotation matrices $(U_i,V_i)$, $i=1,\ldots,m$ such that the objective function in (\ref{eq:handeye}) gives the correct matrix $M$. Note that we do not need to find rotation matrices that map exactly to $M$, since the set $P$ of quadratic forms
that are non-negative is closed and thereby quadratic forms close to $r^TMr$ will also be in $P$, but not in $\Sigma$ (cf.\ Figure~\ref{fig:cone_fig}). In practice, we solve
the following minimization problem with gradient descent:
\begin{equation} \label{eq:handeye-opt-for-M}
\begin{array}{lll}
 &\min\limits_{U_i,V_i \in\SO{3}} & || M(\{ (U_i,V_i) \}_{i=1}^m) -M||_F^2,
\end{array}
\end{equation}
where $M(\{ (U_i,V_i) \}_{i=1}^m)$ is the matrix obtained from formula (\ref{eq:M-handeye}) using $(U_i,V_i)$, $i=1,\ldots,m$ and $M$ is the matrix obtained from Procedure~\ref{proc:P_not_Sigma}. Empirically, we have found that for $m=8$, one can find such rotation matrices which result in a non-tight objective function. 

In the case of registration, we have empirically found
non-tight problem instances
with $5$ point-line correspondences
where the residuals are of the form (\ref{eq:registration}). Similarly to the case of hand-eye calibration in (\ref{eq:handeye-opt-for-M}), we explicitly optimize for point-line correspondences that produce an objective function which is close to the given matrix $M$.

\paragraph{Remark. }
Although Procedure~\ref{proc:P_not_Sigma} in general is guaranteed to find a polynomial $p \in P \setminus \Sigma$, it is not evident {\em a priori} that there exist such polynomials with the $M \succeq 0$ constraint.
Nevertheless, in practice we are able to find many such counterexamples,
with non-tight relaxations.

\newpage
\bibliographystyle{spmpsci}      %
\bibliography{dual}   %

\end{document}